\documentclass[11pt]{article}
\usepackage[final]{acl}
\usepackage{times}
\usepackage{latexsym}
\usepackage[T1]{fontenc}
\usepackage[utf8]{inputenc}
\usepackage{microtype}
\usepackage{inconsolata}
\usepackage{graphicx}
\usepackage{acl}
\title{The Impact of Synthetic Data Augmentation on Discourse-Pragmatic Function Classification}

\author{
 \textbf{Sara Sorahi}$^1$, \textbf{Kevin Tang}$^{2,3}$, \textbf{Reza Kazemian}$^{4}$\\
 Faculty of Arts and Humanities, Heinrich Heine University Düsseldorf \\
 $^1$Institute of Linguistics \hspace{1em} $^2$Department of English Language and Linguistics,\\
 $^3$Department of Linguistics, College of Liberal Arts and Sciences, University of Florida\\
$^4$Sun Yat-sen University, China\\
 \texttt{\{sara.sorahi, kevin.tang\}@hhu.de}, \texttt{reza.kazemian.linguistics@gmail.com}
 }

\begin{document}
\maketitle
\begin{abstract}

Synthetic data augmentation has become a common strategy for
addressing class imbalance in NLP, but most approaches focus
on the \textit{quantity} and \textit{diversity} of generated
examples rather than their geometric relationship to real
training data. We investigate this question in the context of
discourse-pragmatic function classification, a task where data
sparsity is a structural feature rather than a collection
artefact. Using 410 manually annotated instances of the English
word \textit{look} drawn from the British National Corpus,
spanning four functions: Attention Signal, Directive, Discourse
Marker, and Interjection. We generate synthetic training
examples with Llama~3.1 and partition them by their cosine
distance from real training data in RoBERTa embedding space.
We compare six training conditions that differ in the
\textit{placement} of synthetic examples relative to the
empirical decision boundary, while holding augmentation
quantity constant across conditions. All augmented conditions
improve macro-F and accuracy over the real-only baseline, but
core-proximal examples (\textsc{Near}) yield the largest
gains in macro-F ($+0.113$), while a distance-balanced mix
achieves the highest accuracy ($0.748$). No
condition improves AUC, indicating that augmentation shifts
the decision boundary rather than improving the model's
underlying probability estimates. These findings suggest that
\textit{where} synthetic examples land in representation space
matters as much as how many are generated, with implications
for low-resource pragmatic classification more broadly.
\end{abstract}

\section{Introduction}
\label{sec:intro}

Data augmentation has become a standard technique in natural
language processing when labeled data is scarce. Early methods
like lexical perturbation \citep{wei-zou-2019-eda} and
back-translation \citep{sennrich-etal-2016-improving} work by
tweaking the surface form of existing examples while preserving
their meaning. With the rise of large language models, a new
wave of augmentation strategies has emerged, ones that can
generate fluent, label-consistent text on demand
\citep{anaby-tavor-etal-2020-not, brown-etal-2020-language},
and these have delivered strong results across many classification
tasks \citep{feng-etal-2021-survey}.

Most of this work, though, is preoccupied with quantity and
diversity: how many examples to generate, and how varied they
should be. \citet{wang-etal-2025} explicitly optimize for
paraphrase diversity while keeping labels intact, operating on
the assumption that more variety is always better. That assumption
is worth questioning. Synthetic examples that drift too far from
the real data can blur class boundaries, add noise, or shift the
decision boundary in ways that aggregate metrics may not detect.
\citet{sadlier-brown-etal-2024-useful} put this tension in sharp
relief: when they compare model and human performance on natural
versus LLM-generated cloze items, models dramatically outperform
humans on the synthetic ones, a sign that generated text has
its own statistical fingerprint, one that models are good at
exploiting but that may have little to do with genuine linguistic
understanding \citep{bayer-etal-2022-survey}. For
augmentation-based approaches, inflated performance on
synthetic-feeling data is not the same as better pragmatic
classification.

The stakes are even higher when the task involves
discourse-pragmatic function classification, where meaning is
determined by context and communicative intent rather than surface
form. As \citet{ma-etal-2025-pragmatics} document, pragmatic
phenomena remain among the most resistant to computational
treatment, and benchmark resources for evaluating them are still
limited. Matters are made worse by data sparsity: pragmatic
categories like discourse markers and interjections are inherently
low-frequency and hard to accumulate in sufficient numbers
\citep{gries2015}. Even where enough data exists, automatic
disambiguation remains difficult:
\citet{zufferey-popescu-belis-2004-towards} found that classifying
\textit{like} as a discourse marker required careful feature
engineering and still showed limited agreement with human
annotators, illustrating how hard multifunctional pragmatic items
are to classify even under favorable conditions.

Another such multifunctional item, one that has attracted
considerable scholarly attention not only in English but across
its counterparts in other languages, is \textit{look}
\citep{keevalik2008, cardinaletti2015, sanchez-lopez2017,
aijmer2018, nau2021, vanolmen-tantucci-2022,
kazemian-etal-2025b}. It has received
little computational attention, however, and its automatic
function classification remains an open problem. The present
study aims to fill this gap.

Much like \textit{like}, \textit{look} presents a challenge
due to its functional versatility: it can function as an
Attention Signal (AS, e.g., \textit{look, this is important}),
a Directive (DIR, e.g., \textit{look!}), a Discourse Marker
(DM, e.g., \textit{look, I don't think that's right}), or an
Interjection (INTJ, e.g., \textit{oh, look!}). None of these
functions is signaled by the word itself; they emerge from
context, prosody, and pragmatic inference \citep{heine2023},
making \textit{look} both a genuinely hard classification target
and a well-motivated testbed for synthetic augmentation.

We tried several augmentation strategies, starting with
back-translation and moving to prompt-based generation with
Llama~3.1 \citep{grattafiori2024}, served locally via Ollama
\citep{ollama2024}. Rather than treating all synthetic data as
interchangeable, we ask a more precise question: does it matter
\textit{where} in embedding space a synthetic example lands? We
partition generated instances by their cosine distance from real
training data in RoBERTa \citep{liu-etal-2019-roberta} embedding
space and compare how different placement strategies affect
classification performance. Our results show that examples close
to the decision boundary produce the most reliable gains,
suggesting that the effectiveness of augmentation depends not
only on scale or diversity, but on the geometry of representation
space.\footnote{A preliminary version of this study, reporting
only the overall (non-function-level) results as a one-page
abstract, was presented as a poster at KONVENS 2026. This
submission substantially extends that work with the full
experimental analysis, robustness statistics, and function-level
breakdown reported here.}

\section{Data and Annotation}
\label{sec:data}

We focus on the English word \textit{look}, annotated for four
interactive functions following the classification of interactive
grammar \citep{heine2023}. As an \textbf{Attention Signal} (AS),
\textit{look} directs the listener's attention toward what the
speaker is about to say (e.g., \textit{look, this matters}). As
a \textbf{Directive} (DIR), it functions as a frozen stand-alone
expression that retains a residual perceptual meaning without
taking any complement (e.g., \textit{look!}). As a
\textbf{Discourse Marker} (DM), it manages interactional flow
or signals stance (e.g., \textit{look, I don't agree}). As an
\textbf{Interjection} (INTJ), it expresses a spontaneous
emotional reaction (e.g., \textit{oh, look!}). Although all four
functions share the same surface form, they are pragmatically
distinct and require contextual interpretation to disambiguate
\citep{vanolmen-tantucci-2022, landert2023}.

All 410 instances were drawn from the \textbf{British National
Corpus} (BNC) \citep{bnc2007} and manually annotated by two
expert annotators following the parameter-based scheme described
in Appendix~\ref{sec:appendix-annotation}. Annotation focused on
the discourse-pragmatic role of \textit{look} in context rather
than syntactic form alone, since several functions can appear in
structurally similar environments. Ambiguous cases were resolved
through discussion until consensus was reached. As is typical in
discourse-pragmatic annotation \citep{gries2015}, the class
distribution is heavily skewed: AS dominates with 287 instances,
followed by DIR (71), DM (34), and INTJ (18), as shown in
Table~\ref{tab:data}.

\begin{table}[t]
\centering
\small
\begin{tabular}{llcc}
\hline
\textbf{Label} & \textbf{Function} &
\textbf{Real} & \textbf{Augmented} \\
\hline
AS   & Attention Signal  & 287 & 287 \\
DIR  & Directive         & 71  & 287 \\
DM   & Discourse Marker  & 34  & 287 \\
INTJ & Interjection      & 18  & 287 \\
\hline
\textbf{Total} &         & \textbf{410} & \textbf{1,148} \\
\hline
\end{tabular}
\caption{Class distribution before and after augmentation.
AS serves as a real-only reference condition throughout.}
\label{tab:data}
\end{table}

For evaluation, we created five independent stratified 80/20
train--test splits using only attested corpus examples,
preserving the original class distribution in each split. All
reported results are averaged across five runs. Synthetic data
is used only during training; test sets always consist of
authentic BNC examples.

\section{Synthetic Data Augmentation}
\label{sec:augmentation}

Our goal was not simply to generate more data, but to generate
examples that preserved the discourse-pragmatic function of
\textit{look}: a harder problem than it might appear. Unlike
semantic augmentation tasks, where meaning is relatively stable
across paraphrases, pragmatic function is exquisitely sensitive
to context \citep{ma-etal-2025-pragmatics}. A small shift in
wording, register, or surrounding discourse can silently change
how an utterance functions, making standard augmentation methods
unreliable for this kind of data \citep{feng-etal-2021-survey}.

We first explored two translation-based augmentation strategies
before settling on prompt-based LLM generation.

\paragraph{Back-Translation.} Our first attempt used
back-translation \citep{sennrich-etal-2016-improving},
translating existing examples into German\footnote{We experimented with German because it is a high-resource Germanic language like English.} and back into English
to produce paraphrastic variation while preserving meaning. The
most immediate problem was that \textit{look} itself rarely
survived the round-trip. German has no direct pragmatic
equivalent for discourse-functional uses of \textit{look}, so
the translator would routinely drop it, replace it with a
semantically motivated verb, or restructure the sentence
entirely. A sentence like \textit{look, I don't think that's
true}, where \textit{look} is the very item carrying the
discourse-marker function, came back as \textit{however, I
do not believe that to be accurate}: not only had the word \textit{look}
disappeared, but the entire pragmatic character of the utterance
had changed. The conversational informality and the
interactional stance were both missing, and with them the functional
category we were trying to preserve.

To address this, we wrote a small Python script that forced
\textit{look} back into its original position after translation,
reinserting the word at the same slot in the returned sentence.
This kept \textit{look} in place, but introduced a new problem:
the surrounding sentence had already been restructured around
its absence, so the result was often deeply unnatural. A
sentence like \textit{look, we really need to address this now}
might return from German as \textit{we should really address
this now, don't you think}, and after reinsertion become
\textit{look, we should really address this now, don't you
think}: present, but jarring, and no longer clearly a
discourse marker use. The word was there; the pragmatic function
was not.

\paragraph{Cross-Translation.} We then tried cross-translation,
passing examples through two intermediate languages in
sequence, using German and French, hoping that the additional
translation step would introduce more surface variety. It did
introduce variety, but not the kind we needed. Chaining
translations compounded the distortions rather than correcting
them \citep{bayer-etal-2022-survey}. The meaning of sentences
drifted further with each step, and \textit{look} continued to
disappear or mutate. A sentence like \textit{oh, look! They've
already left}, a clear Interjection use, passed through
German as \textit{oh, schau! Sie sind schon gegangen}, then
through French as \textit{oh, regardez! Ils sont déjà partis},
and returned to English as \textit{oh, see! They have already
gone}: the spontaneous affective charge had flattened, the
register had shifted, and \textit{look} had been replaced
altogether. Even when the reinsertion script forced \textit{look}
back in, the surrounding sentence had reorganized itself around
a different word, leaving \textit{look} stranded in a context
that no longer motivated its pragmatic function. The sentence
was grammatical; the category had evaporated.

Both approaches shared the same underlying failure: they treated
\textit{look} as just another content word, when in fact its
pragmatic function is inseparable from the specific interactional
context in which it appears \citep{heine2023}. No amount of
post-hoc reinsertion could recover what the translation process
had already destroyed.

\subsection{Prompt-Based Generation}
\label{subsec:prompting}

We moved to prompt-based generation using \textbf{Llama~3.1 8B}
(Instruct) \citep{grattafiori2024}, served locally via
\textbf{Ollama} \citep{ollama2024} using its default \texttt{Q4\_K\_M}
quantized distribution. A large language model with broad pragmatic
competence can, in principle, generate contextually grounded
examples that preserve discourse-pragmatic function, provided
the prompt is carefully designed \citep{brown-etal-2020-language}.

Early attempts were disappointing. Without sufficient guidance,
the model defaulted to stereotypical constructions: DM examples
invariably opened with \textit{look, the thing is{\ldots}},
directives collapsed into bare imperatives like \textit{look at
this}, and interjections leaned into theatrical exclamatory
phrasing that felt constructed rather than natural. Some outputs
also blurred functional boundaries: \textit{look, just listen
to me}, for instance, could plausibly be read as AS, DM, or DIR
depending on context. This kind of cross-category leakage is a
known risk when generating pragmatically sensitive text without
tight constraints \citep{sadlier-brown-etal-2024-useful}.

We addressed these problems through iterative prompt refinement.
We developed four function-specific prompts, one per category,
each containing a clear functional definition, concrete positive
examples in naturalistic discourse, and explicit instructions
against repetitive openings and formulaic structures. The full
prompts and a representative set of failure cases are documented
in Appendix~\ref{sec:appendix-prompts}. Prompts required
examples to be embedded in plausible conversational or narrative
contexts, since context is precisely what makes pragmatic
function recoverable \citep{ma-etal-2025-pragmatics}. Generation
temperature was set to 1.4 to balance variation against
functional consistency: lower settings produced repetitive
outputs; higher settings caused pragmatic coherence to break
down. All other decoding parameters were left at Ollama's
defaults for this model (top-$k$ = 40, top-$p$ = 0.9, repeat
penalty = 1.1).

The resulting examples were substantially more varied and
contextually grounded. A final-round DM example: \textit{look,
I've been patient, but we really need to make a decision today}
is interactionally motivated and unambiguously a discourse
marker use. A final-round INTJ example: \textit{oh, look!
They've already started without us} is spontaneous, affectively
charged, and clearly distinct from the other categories.

\subsection{Quality Review}
\label{subsec:quality}

All generated outputs were manually reviewed by the same two
annotators responsible for the original corpus annotation, who
independently assessed each example for grammaticality,
naturalness, and functional appropriateness
\citep{wang-etal-2025}. Examples judged ambiguous, repetitive,
pragmatically implausible, or borderline between categories were
discarded. Disagreements were resolved through discussion,
following the same procedure used for the BNC annotation. Only
instances approved by both reviewers were retained, ensuring
that synthetic data met the same basic standards of pragmatic
coherence as the attested corpus examples.

\section{Classification Experiments}
\label{sec:experiments}

\subsection{Sentence Representations}
\label{subsec:embeddings}

All sentences were encoded using \textbf{RoBERTa-base}
\citep{liu-etal-2019-roberta}, a transformer-based language model
pre-trained on large amounts of English text. We used mean pooling
over the final hidden states to produce a single fixed-length
sentence embedding for each instance. The RoBERTa encoder was kept
frozen throughout all experiments; we did not fine-tune it on
the discourse-pragmatic classification task. This design choice
was deliberate: by holding the representation fixed, we ensure
that any differences in classification performance across
conditions can be attributed to the augmentation strategy rather
than to changes in the underlying feature space.
Once all synthetic examples had been encoded, we computed the
distance between each synthetic instance and the real training
set. Specifically, for each synthetic example, we calculated its
mean cosine distance to all real training examples belonging to
the same class. This gave us a distance score for every synthetic
instance, reflecting how far it sits from the core of its
category in representation space.

We then partitioned the synthetic examples for each minority
class into three equal-sized groups based on this distance score:

\begin{itemize}
    \item \textbf{NEAR}: synthetic examples closest to the real
    training instances, sitting near the dense core of the class
    region.
    \item \textbf{MIDDLE}: synthetic examples at an intermediate
    distance, occupying the space between the class core and its
    boundary.
    \item \textbf{FAR}: synthetic examples furthest from the real
    training instances, sitting at or beyond the natural class
    boundary.
\end{itemize}

This partitioning allowed us to directly compare augmentation
strategies that differ not in quantity (all conditions add
the same number of synthetic examples) but in their
geometric relationship to the empirical decision boundary.

\subsection{Training Conditions}
\label{subsec:conditions}

We compare six training conditions (see Table~\ref{tab:conditions}
in the Appendix for full descriptions). The baseline,
\textsc{Real Only}, is trained exclusively on attested BNC
examples. The remaining five conditions each augment the real
training data with synthetic examples for the three minority
classes (DIR, DM, INTJ), scaling each up to 287 instances to
match the majority class (AS). They differ in \textit{which}
synthetic examples are added: \textsc{Real + Near} uses examples
closest to the real training data in embedding space;
\textsc{Real + Far} uses those furthest away; \textsc{Real +
Middle} takes those at an intermediate distance; \textsc{Real +
Random} draws a random sample regardless of distance; and
\textsc{Real + Balanced} combines an equal mix of Near, Middle,
and Far examples.

\subsection{Classifier and Evaluation}
\label{subsec:classifier}

A multinomial logistic regression classifier was trained on top
of the frozen RoBERTa embeddings for four-way discourse-pragmatic
function classification. Logistic regression was chosen
deliberately as a simple, interpretable classifier that places
the explanatory burden on the representation and the training
data rather than on model complexity. This makes it well-suited
for our purpose: we are interested in what the augmentation
strategy contributes, not in maximising raw classification
performance through architectural choices.

Evaluation was carried out using five independent stratified
80/20 train--test splits, with the original class distribution
preserved in each split. All test sets consist exclusively of
authentic BNC examples; synthetic data is never used for
evaluation. We report three metrics averaged across the five
splits: \textbf{macro-F}, which weights each class equally and
is therefore sensitive to minority-class performance;
\textbf{accuracy}, which reflects overall label assignment
correctness; and \textbf{macro-AUC}, which measures the
quality of the model's probability ranking independently of
the final argmax decision. Reporting all three metrics allows
us to distinguish between improvements in hard label
assignment and improvements in the underlying probability
estimates, a distinction that turns out to be central to
interpreting the results.

\section{Results}
\label{sec:results}

\subsection{Overall Performance}
\label{subsec:overall}

\begin{figure}[h]
\centering
\includegraphics[width=\columnwidth]{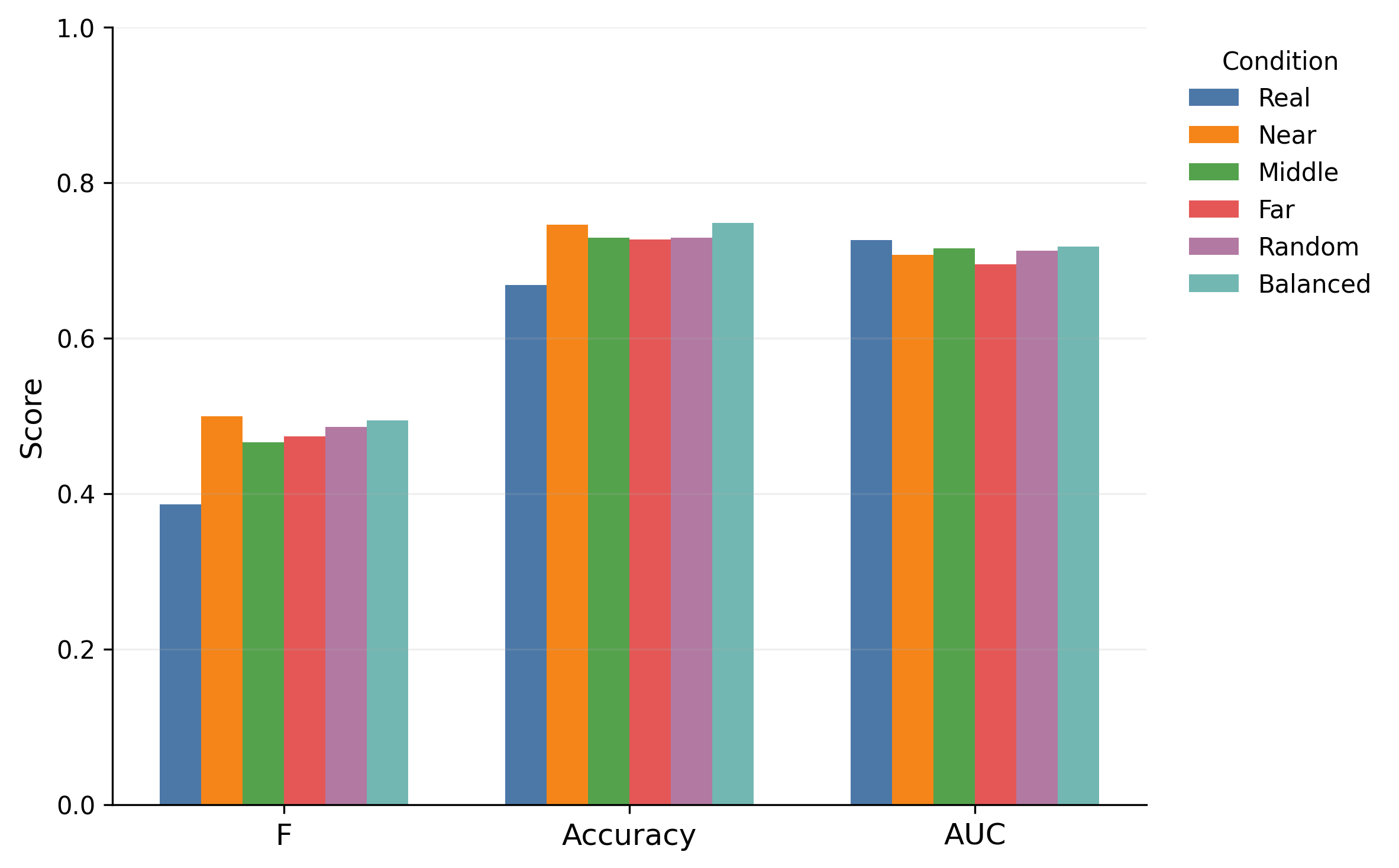}
\caption{Mean macro-F, accuracy, and AUC across six
augmentation conditions, averaged over five stratified
splits. Error bars indicate standard deviation. F and
accuracy improve consistently across all augmented
conditions, while AUC remains flat, consistent with
a decision boundary shift rather than improved probability
estimation.}
\label{fig:barplot}
\end{figure}

Table~\ref{tab:main_results} and Figure~\ref{fig:barplot}
summarise overall classification performance across the six
training conditions, reported as macro-F, accuracy, and AUC
averaged over five stratified train--test splits.

\begin{table*}[t]
\centering
\small
\setlength{\tabcolsep}{12pt}
\begin{tabular}{lccc}
\hline
\textbf{Condition} & \textbf{F} & \textbf{Accuracy} & \textbf{AUC} \\
\hline
Real Only       & 0.386 $\pm$ 0.045 & 0.668 $\pm$ 0.038 & \textbf{0.726 $\pm$ 0.053} \\
Real + Near     & \textbf{0.499 $\pm$ 0.065} & 0.746 $\pm$ 0.033 & 0.707 $\pm$ 0.037 \\
Real + Middle   & 0.466 $\pm$ 0.063 & 0.729 $\pm$ 0.031 & 0.716 $\pm$ 0.023 \\
Real + Far      & 0.474 $\pm$ 0.055 & 0.727 $\pm$ 0.033 & 0.695 $\pm$ 0.034 \\
Real + Random   & 0.486 $\pm$ 0.065 & 0.729 $\pm$ 0.036 & 0.712 $\pm$ 0.044 \\
Real + Balanced & 0.494 $\pm$ 0.051 & \textbf{0.748 $\pm$ 0.040} & 0.718 $\pm$ 0.037 \\
\hline
\end{tabular}
\caption{Overall classification performance across six augmentation
conditions, reported as mean $\pm$ standard deviation over five
stratified train--test splits. Bold indicates the best score per metric.
The real-only baseline achieves the highest AUC despite having the
lowest F and accuracy, a divergence that points to a boundary-shift
effect rather than genuine improvement in probability ranking.}
\label{tab:main_results}
\end{table*}

The results tell a clear and consistent story. Every
augmentation strategy outperforms the real-only baseline on
both macro-F and accuracy, without exception. The largest
gains come from core-proximal augmentation: the
\textsc{Near} condition reaches macro-F of
$0.499 \pm 0.065$, an absolute improvement of $+0.113$
over the baseline ($0.386 \pm 0.045$). Accuracy follows
the same pattern, rising from $0.668 \pm 0.038$ to
$0.746 \pm 0.033$ under the \textsc{Near} condition.
The \textsc{Real+Balanced} condition performs
comparably, achieving the highest mean accuracy overall
($0.748 \pm 0.040$) while its macro-F ($0.494 \pm 0.051$)
falls just short of \textsc{Near}. \textsc{Random},
\textsc{Far}, and \textsc{Middle} all improve over the
baseline too, though by smaller margins, and the ranking
broadly reflects a distance-sensitive pattern: the closer
synthetic examples sit to the natural class boundary, the
more they contribute to classification performance.

AUC tells a strikingly different story. It barely moves
across conditions, ranging from $0.695$ to $0.726$, and
the real-only baseline actually achieves the highest AUC
of all ($0.726 \pm 0.053$). This divergence between AUC
and the other two metrics is theoretically informative.
Because AUC measures how well the model ranks predicted
probabilities, while macro-F and accuracy reflect the
final argmax label assignment, the flat AUC indicates
that augmentation does not improve the model's underlying
confidence estimates; it shifts the decision boundary
so that borderline minority-class instances end up on
the correct side. The model is not becoming more certain;
it is becoming better calibrated at the margins.

\subsection{Robustness and Statistical Significance}
\label{subsec:robustness}

\begin{figure}[h]
\centering
\includegraphics[width=\columnwidth]{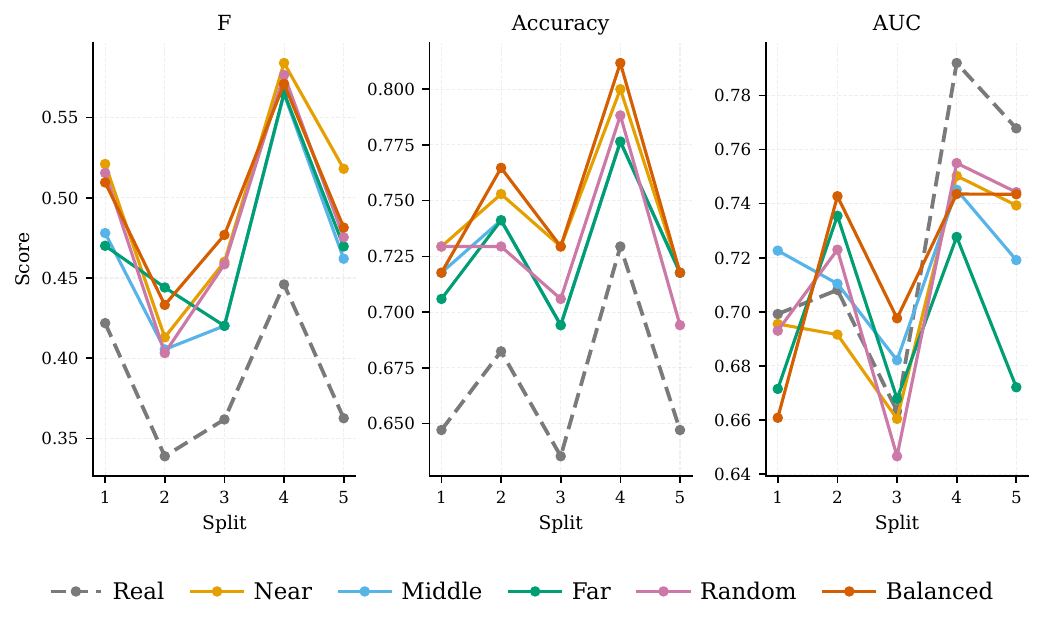}
\caption{Split-wise performance trajectories for macro-F,
accuracy, and AUC across five stratified train--test splits.
Augmented conditions maintain a consistent separation above
the real-only baseline (grey) for F and accuracy across all
five splits. AUC shows no reliable improvement pattern.
The shared performance peak at split 4 suggests this
partition contains a particularly favourable distribution
of boundary cases.}
\label{fig:robustness}
\end{figure}

Paired statistical comparisons between each augmentation
condition and the real-only baseline are reported in
Table~\ref{tab:significance} (Appendix~\ref{sec:appendix-results}).
Figure~\ref{fig:robustness} visualises split-level performance
trajectories for all three metrics.

The most striking finding is the sheer consistency of the
gains. Every augmentation strategy beats the real-only
baseline on both macro-F and accuracy in all five out of
five splits, a result that holds regardless of how the
data is partitioned. This rules out the possibility that
improvements are driven by a single lucky split.

\textsc{Near}-core augmentation remains the strongest
condition for macro-F, with a mean improvement of $+0.113$
($p = 0.0016$, $d = 3.43$). The \textsc{Real+Balanced}
condition is close behind on F ($+0.108$, $p = 0.0001$,
$d = 6.66$) and produces the largest accuracy gain of all
($+0.080$, $d = 8.13$), suggesting that distributing
synthetic examples across distance bands is a reliable
alternative to pure core-proximal sampling. The
remaining conditions, \textsc{Random}, \textsc{Far},
and \textsc{Middle}, all show significant improvements
with large effect sizes, though their gains are somewhat
smaller. \textsc{Middle} produces the weakest F
improvement ($+0.080$) despite sitting between \textsc{Near}
and \textsc{Far} in distance terms, hinting that the
relationship between distance and utility is not strictly
linear.

AUC, by contrast, shows no significant improvement under
any condition. All augmented settings produce small negative
mean differences, none reach statistical significance, and
split-level wins range from just 0/5 to 3/5. This is fully
consistent with the boundary-shift interpretation:
augmentation moves the decision boundary without
meaningfully changing the probability ranking.

Figure~\ref{fig:robustness} makes the split-level picture
concrete. The augmented conditions track closely together
and maintain a clear gap above the real-only baseline for
both F and accuracy across all five splits. The relative
ranking of strategies is stable even as absolute scores
fluctuate. Split 4 stands out as a performance peak for
all conditions, likely because this partition places more
ambiguous boundary cases in the training set, where
boundary-proximal synthetic examples can do the most work.

\subsection{Function-Level Analysis}
\label{subsec:functional}

\begin{figure*}[t]
\centering
\includegraphics[width=0.8\textwidth]{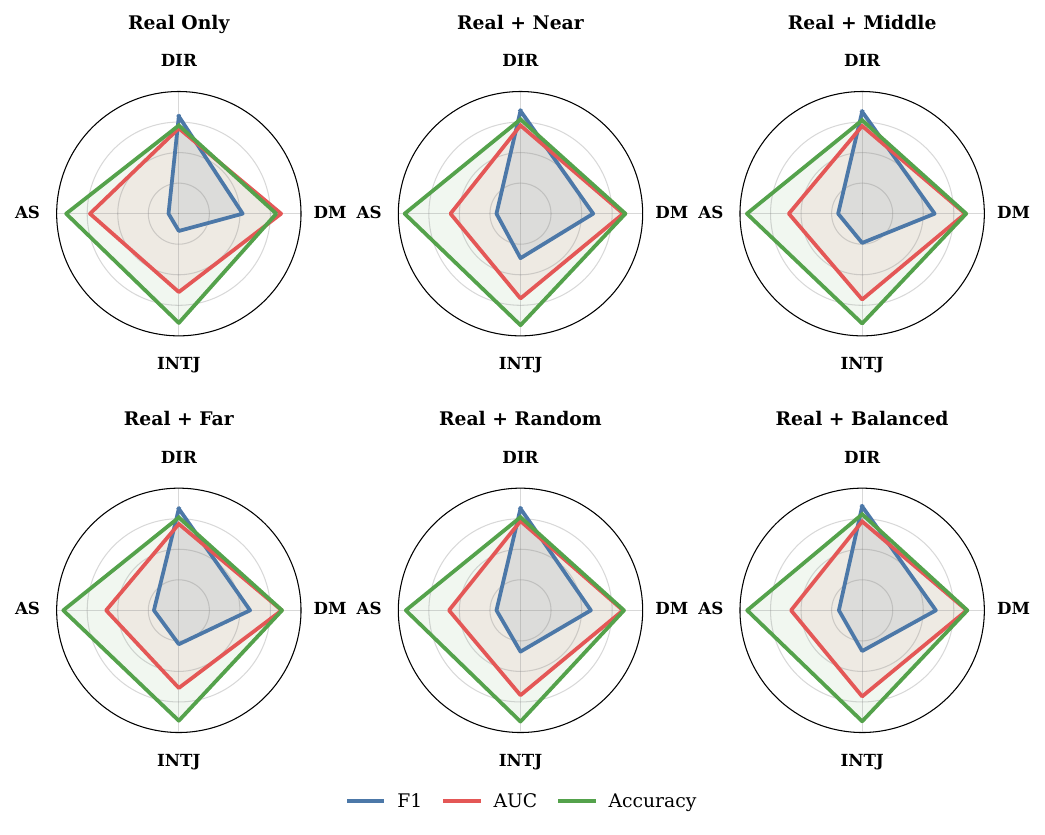}
\caption{Per-condition radar plots showing function-level
F1, AUC, and accuracy across the four
discourse-pragmatic functions (DIR, AS, INTJ, DM). Each plot
corresponds to one training condition. Full numeric values
for all three metrics are reported in
Table~\ref{tab:functional-detailed} (Appendix~\ref{sec:appendix-results}).
The collapsed F1 profile in the Real Only condition,
particularly for INTJ, expands under
augmentation, most notably under \textsc{Near} and
\textsc{Random}. AUC and accuracy profiles stay
stable across conditions, consistent with the boundary-shift
interpretation from Section~\ref{subsec:overall}.}
\label{fig:radar}
\end{figure*}

The overall gains reported above are not distributed evenly
across the four discourse-pragmatic functions. Table~\ref{tab:functional}
reports function-level F1 scores across all six conditions,
and Figure~\ref{fig:radar} shows the full per-function profile
of F1, AUC, and accuracy for each condition as radar plots.
The pattern tracks the degree of class imbalance in the
original data fairly closely: the less real data a category
has, the more it tends to benefit from augmentation.

\begin{table}[t]
\centering
\small
\setlength{\tabcolsep}{10pt}
\begin{tabular}{lcccc}
\hline
\textbf{Condition} & \textbf{DIR} & \textbf{AS} &
\textbf{INTJ} & \textbf{DM} \\
\hline
Real Only       & 0.80 & 0.14 & 0.14 & 0.52 \\
Real + Near     & 0.84 & \textbf{0.20} & \textbf{0.36} & 0.59 \\
Real + Middle   & 0.84 & \textbf{0.20} & 0.24 & 0.59 \\
Real + Far      & 0.83 & \textbf{0.20} & 0.28 & 0.58 \\
Real + Random   & 0.83 & \textbf{0.20} & 0.34 & 0.58 \\
Real + Balanced & \textbf{0.85} & 0.19 & 0.33 & \textbf{0.60} \\
\hline
\end{tabular}
\caption{Function-level F1 scores across augmentation
conditions, averaged over five stratified splits. Bold marks
the best score per function. INTJ and DM benefit most from
augmentation, while AS, the majority class that receives no
synthetic examples, changes little across conditions.}
\label{tab:functional}
\end{table}

Three patterns stand out. First, \textbf{INTJ}, with only 18
real instances the most severely under-represented class,
benefits the most from augmentation. Its F1 score nearly
triples under \textsc{Near}, rising from $0.14$ in the
real-only setting to $0.36$, a gain of $+0.22$ points. That is
a striking jump for a category the real-only classifier can
barely tell apart from noise. Even the weakest strategy,
\textsc{Middle}, still lifts INTJ F1 to $0.24$. The fact that
\textsc{Near} gives the largest INTJ gain while \textsc{Middle}
gives the smallest suggests that synthetic examples close to
the real class core are the most useful
for a category that otherwise has almost no footprint in the
training data.

Second, \textbf{DM} shows reliable but more modest gains,
moving from $0.52$ at baseline to somewhere between $0.58$ and
$0.60$ across augmented conditions, roughly $+0.06$ to $+0.08$
points. These gains hold across all strategies, with
\textsc{Real+Balanced} reaching the highest DM F1 ($0.60$).
For a category with 34 real instances, where the synthetic
examples land matters somewhat, but simply having more
training data to work with matters more.

Third, \textbf{DIR} improves modestly across the board, from
$0.80$ at baseline to between $0.83$ and $0.85$. This smaller
gain fits with DIR already having decent real-data coverage
($n = 71$): the classifier already has enough genuine examples
to learn the category's basic contours, so synthetic data
refines the boundary rather than making the category learnable
in the first place. \textsc{Real+Balanced} gives the best DIR
F1 ($0.85$), suggesting a mix of near and far examples helps
most once a class is already partly learnable.

\textbf{AS}, the majority class with 287 real instances and no
synthetic augmentation, stays largely unchanged across
conditions. Its low F1 ($0.14$--$0.20$) despite being the
dominant class reflects a familiar pathology of imbalanced
classification: the model defaults to AS too often, inflating
recall while hurting precision, which drags F1 down even
though raw accuracy looks fine. Augmentation neither worsens
nor fixes this: AS gets no synthetic examples, so its F1
profile stays essentially flat throughout.

Taken together, the function-level results reinforce the main
finding: augmentation helps most where real data is scarcest,
and the size of the improvement tracks how severely
under-represented a class is. The amount of synthetic data
relative to a class's real-data coverage looks like the main
driver here, with boundary proximity adding a smaller, secondary
benefit, most visible in the INTJ gains under \textsc{Near}
versus \textsc{Middle}.

\section{Conclusion}
\label{sec:conclusion}

We have presented a study of synthetic data augmentation
for discourse-pragmatic function classification, using the
English word \textit{look} as a testbed. Our central
finding is that the effectiveness of augmentation depends
not only on how many synthetic examples are generated,
but on where they sit in embedding space relative to the
real training data. Core-proximal examples, those
closest to the dense center of the real class distribution
in RoBERTa embedding space, produce the largest gains on
macro-F, while a balanced mix of near and far examples
achieves the highest accuracy. None of these
improvements translates to gains in AUC, indicating that
augmentation
shifts the decision boundary rather than improving the
model's underlying probability estimates.

These results suggest that the geometry of synthetic data
in representation space is a meaningful variable that should
be taken into account when designing augmentation strategies
for low-resource pragmatic classification tasks. Simply
generating more examples is not sufficient; where those
examples land matters. We hope this finding motivates
further work on distance- and geometry-aware augmentation
in NLP, particularly for tasks where data sparsity is a
structural feature of the phenomenon rather than a
collection artefact.

\section{Limitations}

Several limitations of the present study should be noted.
First, our corpus is small by NLP standards: 410 attested
instances across four classes, with the two rarest categories
containing only 34 (DM) and 18 (INTJ) real examples. While
this reflects the genuine scarcity of these functions in
naturally occurring speech \citep{gries2015}, it means that
even after augmentation the training sets remain limited, and
the reported improvements should be interpreted with that
context in mind.

Second, the RoBERTa encoder was kept frozen throughout all
experiments. This was a deliberate design choice to isolate
the contribution of augmentation strategy from changes in
the underlying feature space, but it also means that the
representations used for classification were not adapted to
the distributional properties of discourse-pragmatic data.
Fine-tuning the encoder on the full augmented training set
may yield further gains, and the interaction between
encoder adaptation and distance-based augmentation remains
an open question. Relatedly, mean-pooled representations from
frozen transformer encoders like RoBERTa are known to exhibit
anisotropy \citep{ethayarajh-2019-contextual}, which can distort
cosine-based distance measures; it remains an open question
whether our distance-sensitive findings would hold under
embeddings specifically designed for cosine geometry, such as
sentence-transformers \citep{reimers-gurevych-2019-sentence}.

Third, our distance-based partitioning relies on mean cosine
distance to the real training set within each class. This is
a simple and interpretable measure, but it does not capture
the full geometry of the embedding space, for instance,
whether a synthetic example sits in a dense or sparse region,
or how close it is to examples from \textit{other} classes.
More sophisticated geometric criteria may further refine the
augmentation strategy.

Fourth, our statistical comparisons (Appendix~\ref{sec:appendix-results},
Table~\ref{tab:significance}) are computed over only five
stratified splits. With $n = 5$, p-values and effect sizes are
sensitive to the particular partitions drawn, and we do not
report confidence intervals or bootstrap estimates that would
better characterize the uncertainty around these differences.
The consistency of the ranking across all five splits partially
mitigates this concern, but a larger number of splits or a
bootstrap-based reanalysis would strengthen the statistical
claims.

Fifth, our experiments are conducted on a single language
(English) and a single word (\textit{look}). Whether the
findings generalise to other polyfunctional items, other
languages, or other discourse-pragmatic classification tasks
remains to be established. The cross-linguistic literature
on \textit{look}
\citep{keevalik2008, vanolmen-tantucci-2022} suggests that
similar functional ambiguity arises across languages, but the
degree to which synthetic augmentation addresses it may vary.

Sixth, our prompt templates impose different structural
constraints across functions: INTJ examples, for instance, were
generated as three-sentence mini-scenes (setup, target sentence,
reaction), while DM and DIR examples follow shorter templates.
We did not verify that attested BNC instances of each function
are matched in length or discourse structure to their synthetic
counterparts. If real and synthetic examples differ systematically
in context length by function, this could confound the
distance-based effects we report with a simpler input-length
effect, and this possibility warrants further investigation.

Finally, we note that the prompt-based generation approach
depends on the quality of the underlying language model
(Llama~3.1) and the careful manual design of prompts. Our
results may not transfer directly to other models or domains
without similar iterative refinement of the generation
constraints.

\section{Ethical Considerations}

The data used in this study are drawn from the British
National Corpus (BNC) \citep{bnc2007}, a publicly available
resource distributed under an established licence for
academic research. No new human subjects data were collected,
and no personally identifiable information is present in the
corpus examples used for annotation or evaluation.

The synthetic data generated in this study were produced
using Llama~3.1 \citep{grattafiori2024}, a publicly released
model, and served locally via Ollama \citep{ollama2024}.
No proprietary APIs or closed systems were used in the data
generation pipeline. Generated examples were manually reviewed
for appropriateness before inclusion in the training set.

The annotation work was carried out by two expert annotators
with a background in linguistics. No crowd-sourced or
low-paid annotation labour was used, and no vulnerable
populations were involved in any part of the study.

We do not foresee significant direct harms arising from this
research. The task, automatic classification of
discourse-pragmatic functions of a single word, is a basic
NLP research problem without immediate dual-use risks.
However, we acknowledge that improvements in pragmatic
function classification could in principle contribute to
downstream applications in dialogue systems or automated
discourse analysis, and that such applications should be
developed and deployed with appropriate scrutiny.

\bibliography{custom}
\clearpage
\appendix

\section{Annotation Details}
\label{sec:appendix-annotation}

This appendix provides the annotation guidelines used for
coding \textit{look} as an interactive element in spoken
discourse \citep{heine2023}. Our framework builds on the
cross-linguistic parameters introduced by
\citet{vanolmen-tantucci-2022} for the study of parenthetical
\textit{look}, extended with two additional parameters,
argument structure and co-occurrence, following
\citet{heine2023}. We distinguish four primary functions:
Directive (DIR), Attention Signal (AS), Discourse Marker (DM),
and Interjection (INTJ). Annotation covers five parameters per
instance: turn position, clause position, perceptual meaning,
speech act, and argument structure.

\begin{table*}[t]
\centering
\small
\begin{tabular}{llllll}
\hline
\textbf{Function} & \textbf{Bleached} & \textbf{Speech act} &
\textbf{Arg. structure} & \textbf{Clause pos.} &
\textbf{Turn pos.} \\
\hline
Directive        & No               & Directive  & S, H          & Initial          & Initial \\
Attention Signal & Partially        & Directive  & S, H, T       & Initial          & Initial \\
Discourse Marker & Yes              & ---        & T1, S, (H), T2 & Initial/Medial  & Initial/Medial \\
Interjection     & Yes              & Expressive & S             & Initial/Med./Final & Initial/Med./Final \\
\hline
\end{tabular}
\caption{Annotation scheme: parameters and values by function.}
\label{tab:annotation-scheme}
\end{table*}

\paragraph{Turn position / Co-occurrence.}
Turn position refers to the placement of \textit{look} within
a speaker's turn: \textbf{initial}, \textbf{medial}, or
\textbf{final}. Instances appearing at the start of quotations
or reported speech are treated as turn-initial.

\paragraph{Clause position.}
Clause position follows the same three values (initial, medial,
final). Position is a key disambiguating cue: initial position
correlates with linking to preceding discourse and topicalizing,
while final position anticipates upcoming discourse
\citep{heine2023}.

\paragraph{Perceptual meaning.}
This parameter captures whether \textit{look} has retained its
original lexical meaning or undergone pragmatic bleaching
\citep{fagard2010, heine2023, kazemian-etal-2025b}. Values:
\textit{Yes} (fully preserved, e.g.\ directive use);
\textit{To some extent} (partially preserved, e.g.\ attention
signal); \textit{No} (fully bleached, e.g.\ discourse marker,
interjection).

\paragraph{Speech act.}
Following \citet{searle1979}, we distinguish two speech acts
performed by \textit{look} itself: \textit{directive} (DIR and
AS, where the speaker prompts the hearer to act or attend) and
\textit{expressive} (INTJ, where the speaker expresses a
psychological state). When neither is present, \textit{look}
functions as a DM or AS \citep{vanolmen-tantucci-2022}.

\paragraph{Argument structure.}
Following \citet{heine2023}, we annotate argument structure
valence: \textbf{S} (speaker), \textbf{H} (hearer), and
\textbf{T} (theme: the state of affairs prompting the
interactive). Interjections involve S only; directives involve
S and H; discourse markers involve T1, S, (H), and T2.

\paragraph{Annotator roles.}
A \textit{primary annotator} identified occurrences and assigned
all parameter values. A \textit{secondary annotator} reviewed
annotations and participated in conflict resolution. Discrepancies
were resolved through discussion until consensus was reached.

\vspace{8pt}
\noindent The following example illustrates the annotation
process on a naturally occurring BNC excerpt.

\vspace{4pt}
\noindent\textit{A: It becomes more and more apparent that the
unassuming, virtually ego-free Mr.\ Cooder is far happier
discussing musicians other than himself\ldots}\\
\textit{B: Well, \textbf{look}, I mean, I just play.}

\begin{table*}[t]
\centering
\small
\begin{tabular}{lllllll}
\hline
\textbf{Token} & \textbf{Function} & \textbf{Turn pos.} &
\textbf{Clause pos.} & \textbf{Bleached} &
\textbf{Speech act} & \textbf{Arg. structure} \\
\hline
\textit{look} & DM & Initial & Initial & Yes & --- & (T1, S, H, T2) \\
\hline
\end{tabular}
\caption{Worked annotation example.}
\label{tab:worked-example}
\end{table*}

\noindent\textit{Justification:} \textit{Look} appears
turn-initially and clause-initially. It is fully bleached,
performs no directive or expressive speech act, and manages
discourse coherence by marking a shift in the conversation.
The argument structure involves T1 (the preceding speaker's
belief), S (speaker B), H (speaker A), and T2 (the current
situation being introduced).

\section{Prompts}
\label{sec:appendix-prompts}

\subsection{Prompt Iteration: Failure Cases}
\label{subsec:example-outputs}

\begin{table*}[t]
\centering
\small
\begin{tabular}{p{0.30\textwidth} p{0.30\textwidth} p{0.30\textwidth}}
\hline
\textbf{Generated output} & \textbf{Violation} & \textbf{Fix applied} \\
\hline
``She wanted to \textit{look for} her keys before leaving.''
&
Forbidden phrasal verb (\textit{look for}) generated under the DM condition; \textit{look} retains its literal search sense rather than functioning as a discourse marker.
&
Added an explicit list of forbidden phrasal patterns and a final self-check instruction requiring the model to verify their absence before output.
\\
\hline
``\textit{Look!} I can't believe you did that, that's amazing!''
&
Generated under the DM condition but functions as an emotional exclamation (INTJ), with \textit{look} expressing surprise rather than organizing discourse.
&
Added an explicit negative definition (``It does NOT express perception, emotion, or command'') to the DM prompt to block cross-category leakage.
\\
\hline
``You should \textit{look} at this before deciding.''
&
Generated under the DIR condition but uses a modal construction (\textit{should}) and the phrasal verb \textit{look at}, rather than a bare imperative.
&
Restricted DIR position to Initial-only and required a bare imperative verb immediately following \textit{Look,}, explicitly banning modal/declarative forms.
\\
\hline
``That's, look, the third time this week, look, that this has happened.''
&
Two instances of \textit{look} in a single example, generated under the INTJ condition.
&
Added an explicit requirement that each example contain exactly one instance of \textit{look}/\textit{Look}.
\\
\hline
``Look out, the meeting is about to start.''
&
Forbidden phrasal verb (\textit{look out}) generated under the DIR condition; reads as a warning rather than a directive introducing an imperative.
&
Added \textit{look out} to the forbidden-pattern list and required the imperative verb to be a concrete bare verb (e.g., stop, come, take) rather than a continuation of \textit{look}.
\\
\hline
\end{tabular}
\caption{Constraint violations observed during early prompt iterations.}
\label{tab:failure-cases}
\end{table*}

\subsection{Final Prompt Templates}
\label{subsec:prompt-templates}

\begin{table*}[t]
\centering
\small
\begin{tabular}{p{0.95\textwidth}}
\hline
\textbf{Prompt: Discourse Marker (DM)} \\
\hline
You are generating exactly N English examples that use the word ``look'' or ``Look'' (not looks, looking, or looked).
\newline\textbf{FUNCTION:} Discourse Marker (DM). ``Look'' functions as a discourse marker: a pragmatic particle that organizes and structures discourse. It introduces clarification, justification, contrast, reformulation, or stance. It does NOT express perception, emotion, or command. It must sound natural in spoken or conversational written English.
\newline\textbf{POSITION:} Initial or Medial only (never final).
\newline\textbf{FORBIDDEN PATTERNS:} ``look at'', ``look out'', ``look over'', ``look around'', ``look into'', ``look for'', ``look this way''; any perceptual or emotional sense; ``Look!'' interjections; directives (e.g., ``Look, stop talking.'').
\newline\textbf{OUTPUT FORMAT:} Produce exactly N numbered examples with three lines each: \textit{Before:}, \textit{DM:}, \textit{After:}, separated by a blank line.
\\
\hline
\end{tabular}
\caption{Prompt template for generating synthetic examples of \textit{look} as a discourse marker (DM).}
\label{tab:prompt-dm}
\end{table*}

\begin{table*}[t]
\centering
\small
\begin{tabular}{p{0.95\textwidth}}
\hline
\textbf{Prompt: Interjection (INTJ)} \\
\hline
You are generating exactly N English examples where the word ``look'' or ``Look'' (not looks, looking, or looked) functions purely as an interjection (INTJ).
\newline\textbf{FUNCTION:} Interjection (INTJ). ``Look'' expresses spontaneous emotion, such as surprise, admiration, joy, excitement, frustration, disbelief, or shock. It does not mean see, watch, observe, inspect, or involve any literal act of perception.
\newline\textbf{REQUIRED PROPERTIES:} Exactly one ``look''/``Look'' per example; balanced mix of Initial, Medial, and Final positions (about one-third each).
\newline\textbf{OUTPUT FORMAT:} Produce exactly N numbered examples, each a coherent three-sentence mini-scene (setup, target sentence, reaction), with no line labels.
\\
\hline
\end{tabular}
\caption{Prompt template for generating synthetic examples of \textit{look} as an interjection (INTJ).}
\label{tab:prompt-intj}
\end{table*}

\begin{table*}[t]
\centering
\small
\begin{tabular}{p{0.95\textwidth}}
\hline
\textbf{Prompt: Directive (DIR)} \\
\hline
\textbf{FUNCTION:} Directive (DIR). ``Look'' functions as a directive discourse particle introducing a true imperative: a direct command, request, or instruction that urges the hearer to act immediately. It marks insistence or urgency before an imperative clause and must not refer to perception or visual direction.
\newline\textbf{POSITION:} Initial only. Each imperative must contain a bare verb (e.g., stop, come, take, listen, tell, move, wait, help, write, sit, give). No modal or declarative constructions.
\newline\textbf{OUTPUT FORMAT:} Produce exactly N numbered examples, each three sentences (setup, ``Look,'' + imperative, reaction/result), with no line labels.
\\
\hline
\end{tabular}
\caption{Prompt template for generating synthetic examples of \textit{look} as a directive (DIR).}
\label{tab:prompt-dir}
\end{table*}

\section{Supplementary Results and Analysis}
\label{sec:appendix-results}

\begin{table*}[t]
\centering
\small
\begin{tabular}{lp{12cm}}
\hline
\textbf{Condition} & \textbf{Description} \\
\hline
\textsc{Real Only} & Trained on attested BNC examples only; serves as the baseline. No synthetic data is added. \\
\textsc{Real + Near} & Real data augmented with synthetic examples whose mean cosine distance to real training instances of the same class is smallest (closest to the dense core of the class region). \\
\textsc{Real + Middle} & Real data augmented with synthetic examples at an intermediate distance from the real training set, occupying the space between the class core and its boundary. \\
\textsc{Real + Far} & Real data augmented with synthetic examples furthest from the real training instances, sitting at or beyond the natural class boundary in embedding space. \\
\textsc{Real + Random} & Real data augmented with a random sample of synthetic examples drawn without regard to distance from the real training set. \\
\textsc{Real + Balanced} & Real data augmented with an equal mix of Near, Middle, and Far synthetic examples, distributing augmentation evenly across the full range of embedding distances. \\
\hline
\end{tabular}
\caption{Full descriptions of the six training conditions.}
\label{tab:conditions}
\end{table*}

\begin{table*}[t]
\centering
\small
\setlength{\tabcolsep}{6pt}
\begin{tabular}{llcccc}
\hline
\textbf{Condition} & \textbf{Metric} & $\Delta$ &
\textbf{\textit{p}} & \textbf{\textit{d}} & \textbf{Wins} \\
\hline
Near     & F        & +0.113 & 0.0016 & 3.43 & 5/5 \\
Near     & Accuracy & +0.078 & 0.0001 & 7.38 & 5/5 \\
Near     & AUC      & $-$0.019 & 0.066 & $-$1.12 & 0/5 \\
\hline
Middle   & F        & +0.080 & 0.003  & 2.87 & 5/5 \\
Middle   & Accuracy & +0.061 & 0.0002 & 6.22 & 5/5 \\
Middle   & AUC      & $-$0.010 & 0.553 & $-$0.29 & 3/5 \\
\hline
Far      & F        & +0.088 & 0.004  & 2.73 & 5/5 \\
Far      & Accuracy & +0.059 & 0.0001 & 7.07 & 5/5 \\
Far      & AUC      & $-$0.031 & 0.236 & $-$0.62 & 2/5 \\
\hline
Random   & F        & +0.100 & 0.0008 & 4.05 & 5/5 \\
Random   & Accuracy & +0.061 & 0.0009 & 3.99 & 5/5 \\
Random   & AUC      & $-$0.014 & 0.191 & $-$0.70 & 1/5 \\
\hline
Balanced & F        & +0.108 & 0.0001 & 6.66 & 5/5 \\
Balanced & Accuracy & +0.080 & 0.0001 & 8.13 & 5/5 \\
Balanced & AUC      & $-$0.008 & 0.663 & $-$0.21 & 2/5 \\
\hline
\end{tabular}
\caption{Paired statistical comparison of each augmentation condition against the real-only baseline.}
\label{tab:significance}
\end{table*}

\begin{table*}[t]
\centering
\small
\setlength{\tabcolsep}{5pt}
\begin{tabular}{lccc ccc ccc ccc}
\hline
& \multicolumn{3}{c}{\textbf{DIR}} & \multicolumn{3}{c}{\textbf{AS}} &
\multicolumn{3}{c}{\textbf{INTJ}} & \multicolumn{3}{c}{\textbf{DM}} \\
\cline{2-4} \cline{5-7} \cline{8-10} \cline{11-13}
\textbf{Condition} & F1 & AUC & Acc & F1 & AUC & Acc & F1 & AUC & Acc & F1 & AUC & Acc \\
\hline
Real Only       & 0.80 & 0.73 & 0.67 & 0.14 & 0.73 & 0.67 & 0.14 & 0.73 & 0.67 & 0.52 & 0.73 & 0.67 \\
Real + Near     & 0.84 & 0.71 & 0.75 & 0.20 & 0.71 & 0.75 & 0.36 & 0.71 & 0.75 & 0.59 & 0.71 & 0.75 \\
Real + Middle   & 0.84 & 0.72 & 0.73 & 0.20 & 0.72 & 0.73 & 0.24 & 0.72 & 0.73 & 0.59 & 0.72 & 0.73 \\
Real + Far      & 0.83 & 0.70 & 0.73 & 0.20 & 0.70 & 0.73 & 0.28 & 0.70 & 0.73 & 0.58 & 0.70 & 0.73 \\
Real + Random   & 0.83 & 0.71 & 0.73 & 0.20 & 0.71 & 0.73 & 0.34 & 0.71 & 0.73 & 0.58 & 0.71 & 0.73 \\
Real + Balanced & 0.85 & 0.72 & 0.75 & 0.19 & 0.72 & 0.75 & 0.33 & 0.72 & 0.75 & 0.60 & 0.72 & 0.75 \\
\hline
\end{tabular}
\caption{Function-level F1, plus overall (macro-averaged) AUC and
accuracy, for all six augmentation conditions. F1 is computed
per function; AUC and accuracy are macro-averaged classifier-level
metrics (identical to Table~\ref{tab:main_results}) and are
repeated across each condition's row, not computed separately
per function, since AUC and accuracy are not well-defined for a
single class in isolation. Only the F1 columns vary within a row.}
\label{tab:functional-detailed}
\end{table*}

\end{document}